\documentclass{article}

\usepackage{microtype}
\usepackage{graphicx}
\usepackage{subfigure}
\usepackage{booktabs} 
\usepackage{hyperref}

\usepackage[accepted]{icml2024}

\usepackage{amsmath}
\usepackage{amssymb}
\usepackage{mathtools}
\usepackage{amsthm}
\usepackage{bbold}
\usepackage[capitalize,noabbrev]{cleveref}

\theoremstyle{plain}

\theoremstyle{definition}

\theoremstyle{remark}

\usepackage{tcolorbox}  

\newcommand{\position}[2]{%
\begin{tcolorbox}[colback=gray!10!white,leftrule=2.5mm,size=title]
\textbf{#1}: #2
\end{tcolorbox}
\vspace{-0.1cm}%
}

\icmltitlerunning{Open-Endedness is Essential for Artificial Superhuman Intelligence}

\begin{document}

\twocolumn[
\icmltitle{Open-Endedness is Essential for Artificial Superhuman Intelligence}

\icmlsetsymbol{equal}{*}

\begin{icmlauthorlist}
\icmlauthor{Edward Hughes}{equal,gdm}
\icmlauthor{Michael Dennis}{equal,gdm}
\icmlauthor{Jack Parker-Holder}{gdm}
\icmlauthor{Feryal Behbahani}{gdm}
\icmlauthor{Aditi Mavalankar}{gdm}
\icmlauthor{Yuge Shi}{gdm}
\icmlauthor{Tom Schaul}{gdm}
\icmlauthor{Tim Rockt\"aschel}{gdm}
\end{icmlauthorlist}

\icmlaffiliation{gdm}{Google DeepMind, London, UK}

\icmlcorrespondingauthor{Edward Hughes}{edwardhughes@google.com}
\icmlcorrespondingauthor{Michael Dennis}{dennismi@google.com}

\icmlkeywords{Open-Endedness, Foundation Models, Artificial Superhuman Intelligence, AI Safety}

\vskip 0.3in
]

\printAffiliationsAndNotice{\icmlEqualContribution}

\begin{abstract}
In recent years there has been a tremendous surge in the general capabilities of AI systems, mainly fuelled by training foundation models on internet-scale data. Nevertheless, the creation of open-ended, ever self-improving AI remains elusive. \textbf{In this position paper, we argue that the ingredients are now in place to achieve \emph{open-endedness} in AI systems with respect to a human observer. Furthermore, we claim that such open-endedness is an essential property of any artificial superhuman intelligence (ASI).} We begin by providing a concrete formal definition of open-endedness through the lens of novelty and learnability. We then illustrate a path towards ASI via open-ended systems built on top of foundation models, capable of making novel, human-relevant discoveries. We conclude by examining the safety implications of generally-capable open-ended AI. We expect that open-ended foundation models will prove to be an increasingly fertile and safety-critical area of research in the near future. 
\end{abstract}

\section{Introduction}

Recent years have seen impressive progress in AI, mainly driven by foundation models~\citep{bommasani2021opportunities}. These models are increasingly used as agents in various applications~\citep[e.g., ][]{wang2023voyager,wu2023spring,lifshitz2023steve1,wang2023jarvis1,liu2023agentbench,zheng2024gpt4v,ahn2022can}. This represents significant progress towards artificial general intelligence~(AGI), in the sense of reaching human-level performance on a wide range of tasks \citep{legg2007universal}. However, we are still missing a formal description of what it would take for an autonomous system to self-improve towards increasingly creative and diverse discoveries \emph{without end}---a Cambrian explosion of emergent capabilities, behaviors, and artifacts. This kind of \textit{open-ended} invention is the mechanism by which human individuals and society at large accumulates new knowledge and technology. Therefore, open-endedness must be a property of an artificial superhuman intelligence \citep[ASI,][]{morris2023levels} that can, by definition, accomplish a wide range of tasks at a level which no human can match. By the very nature of superhuman intelligence, open-ended discovery of innovative solutions is essential to empower humanity to manage its risks, just as society evolves norms and institutions to govern increasingly capable humans across generations \cite{richerson2001institutional}.
 
Foundation models such as large language models (LLMs) have scaled learning to large, static datasets scraped from the internet. Extrapolating, we may soon be running out of high-quality textual and visual data for training such models~\citep{villalobos2022will}. Thus, open-endedness is unlikely to arise for free by training on ever-larger datasets. Rather, a system endowed with the open-endedness necessary for ASI will eventually have to create, refute and refine its own explanatory knowledge, in interaction with a source of evidence \citep{deutsch2011beginning}, as well as learning what data to learn from~\citep{jiang2022general}. Moreover, for ASI to be useful and safe, it is important that open-endedness be guided towards knowledge that is understandable by and beneficial for humanity. Foundation models and open-endedness are orthogonal dimensions, whose combination is particularly powerful~\citep[cf.][]{lehman2022evolution,huang2022large,chen2023evoprompting,meyerson2023language,zhang2023omni,wu2023spring,wang2023voyager}. Open-ended algorithms endow foundation models with the ability to uncover new knowledge, while foundation models guide the search space for open-ended AI towards discovering human-relevant artifacts efficiently \citep{liu2023large, ma2023eureka,romera-paredes2024mathematical}. A formal definition of open-endedness can catalyze progress in this direction, offering clarity and focus to galvanize the research community.  

We provide a new and precise definition of open-endedness in Section \ref{sec:definition}, inspired by the open-ended systems in nature that have created life, the human brain, culture, and technology, as well as open-ended systems in silico that, for instance, have achieved superhuman level at the game of Go~\citep{silver2016mastering}, generated human-level adaptation to novel 3D tasks~\citep{ada}, self-improved language models~\citep{fernando2023promptbreeder, yang2023large}, unlocked the tech tree in Minecraft~\citep{wang2023voyager}, and discovered new results in pure mathematics~\citep{romera-paredes2024mathematical}. Open-endedness has been understood in a wide variety of ways \cite{earle2021video} ever since it gained prominence as a term in the study of artificial life \citep{bedau1992measurement, bedau11998classification} and biological evolution ~\cite{holland1992adaptation,mcshea1996perspective, waddington2008paradigm}. Contrary to~\citet{stepney2023open}, we believe quantifying open-endedness is both possible and important going forward, and, akin to~\citet{sigaud2023definition}, we believe it can be achieved via the help of an observer external to the system. Our definition makes formal the aphorism of Lisa B. Soros that, as observers of an open-ended system, ``we'll be surprised but we'll be surprised in a way that makes sense in retrospect''. Concretely, open-ended systems produce increasingly novel and surprising artifacts that are hard to predict, even for an observer who has learned to better predict by examining past artifacts. Once a system exhibits these characteristics, i.e. producing learnable but novel artifacts, we call it an open-ended system. This allows us to pinpoint the sense in which open-endedness is essential for ASI, to provide examples illustrating how existing open-ended AI systems lack generality, and to argue that present-day foundation models are not yet open-ended. 

Historically, the field of open-endedness has faced numerous challenges. Principal among these has been the problem of structuring the search space so as to regularly produce artifacts which are both novel and interesting to humans~\citep{ma2023eureka}. When humans make discoveries, they do so by ``standing on the shoulders of giant human datasets''~\citep{jeffclunegiants2022}; that is to say, utilising prior world, domain and commonsense knowledge, which they have acquired biologically or culturally. Since foundation models have been trained on vast amounts of human data, they capture human notions of interestingness~\citep{zhang2023omni}. Furthermore, they are general sequence modellers~\citep{mirchandani2023large} and can generate variations from existing examples~\citep{meyerson2023language}, thus serving as general mutation operators. This is compelling since with more advanced foundation models, practical implementations of open-ended systems  become increasingly feasible. Taken together, \textit{open-ended foundation models} can both vary (i.e., mutate) data and assess novelty and interestingness of real and generated data to decide what data to further explore (i.e., select)~\citep{jiang2022general}.

In Section \ref{sec:oe-fm} we provide some concrete research directions for this marriage between open-endedness and foundation models, for example leveraging evolutionary algorithms and reinforcement learning. Generally capable open-ended systems may be both extremely powerful and increasingly prevalent, prompting pressing safety considerations~\citep{ecoffet2020open}. In Section \ref{sec:safety}, we argue that research into open-ended systems will be essential to safely and beneficially deploy any increasingly general and autonomous AI.

\section{Defining Open-Endedness}\label{sec:definition}

\subsection{Formal Definition}\label{sec:formal_definition}

\begin{figure}[t!]
    \centering
    \includegraphics[width=0.95\linewidth]{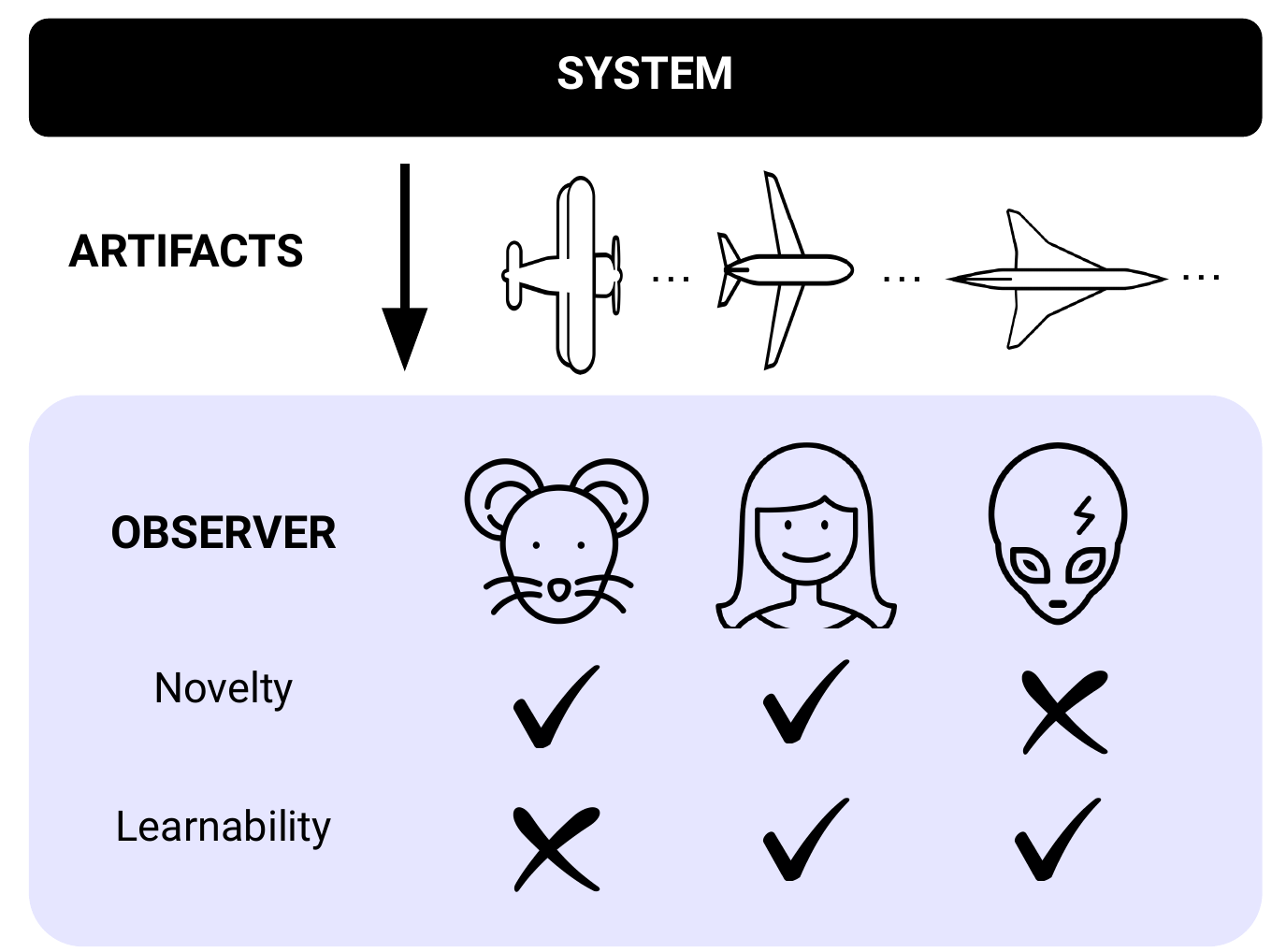}
    \caption{\textbf{Illustration of open-endedness definition.} The definition of open-endedness hinges on a system's ability to continuously generate artifacts that are both novel and learnable to an observer. Consider a system that designs various aircraft: a mouse (left) might find these designs novel but lack the capacity to comprehend the principles behind them; for a human studying aerospace engineering (middle), the system offers both novelty and the potential for learning, making it open-ended. However, a superintelligent alien (right) with vast aerospace knowledge might not find the design novel, but would still be able to analyze and understand them. This highlights that open-endedness is \emph{observer-dependent} and that novelty or learnability alone is not enough.}
    \label{fig:definition}
    \vspace{-8pt}
\end{figure}

The notion of an open-ended system has received many colloquial definitions \citep{chromaria,stanley2015greatness,stanley2017open,why_oe_matters}. More formal approaches have often focused on the case of evolutionary systems, quantifying the increasing complexity~\cite{mcshea1996perspective, waddington2008paradigm} and perpetual novelty~\cite{holland1992adaptation} of biological evolution. Intuitively, an open-ended system endlessly produces novel and interesting artifacts. But novelty and interestingness have generally been characterised without sufficient precision, or in an overly narrow way. We provide a general-purpose, formal definition of open-endedness, as follows.

\position{Definition}{From the perspective of an observer, a system is \textit{open-ended} if and only if the sequence of artifacts it produces is both novel and learnable.}
\vskip 0.05in

More formally, a \textbf{system} $S$ produces a sequence of \textbf{artifacts} $X_t$, indexed by time $t$. An \textbf{observer} $O$ processes a new artifact $X_T$ to determine its predictability given a history $X_{1:t}$ of past ones. $O$ possesses a \textbf{statistical model} $\hat{X}_t$ which predicts an arbitrary future artifact based on its observations of the artifacts it has seen up to time $t$. The observer judges the quality of their prediction based on a \textbf{loss metric} $\ell (\hat{X}_t, X_T)$, or $\ell(t,T)$ for short. A natural implementation of $\hat{X}_t$ is as a learning algorithm. 

A system displays \textbf{novelty} if artifacts become increasingly unpredictable with respect to the observer’s model at any fixed time $t$, namely:
\[
\forall t, \forall T > t, \exists T' > T: \mathbb{E}\left[\ell (t, T')\right] > \mathbb{E}\left[\ell (t, T)\right]\,.
\]
In other words, there is always a less predictable artifact coming further in the future.\footnote{We take the expectation over any stochasticity in the artefacts; practically speaking, were the observer to make observations from identical copies of the system $S$, the expectation of $\ell$ would be approximated by the empirical mean.}

The system is \textbf{learnable} whenever conditioning on a longer history makes artifacts more predictable, namely:
\[
\forall T, \forall t < T, \forall T > t ' > t: \mathbb{E}\left[\ell (t', T)\right] < \mathbb{E}\left[\ell (t, T)\right]\, .
\]
Finally, a system is \textbf{open-ended} from the perspective of the observer $O$ if and only if it generates sequences of artifacts that are both novel and learnable (see Figure \ref{fig:definition}). The novelty aspect ensures the presence of information gain within the system, while learnability guarantees that this information gain holds meaning and is ``interesting'' to the observer. 

For example, imagine that the system is a noisy TV producing uniform random noise \citep{burda2018exploration}. A noisy TV is learnable, allowing the observer to learn a statistical model that approximates the uniform distribution increasingly well; however, once the observer’s model converges to uniform the system loses its novelty: all that is left is aleatoric uncertainty, which is collapsed by the expectation. Now imagine that the system is a noisy TV switched periodically by a remote control to a random, arbitrary distribution. Every time the channel is changed, the observer may experience novelty; however, the system is now not learnable, because the history of artifacts (previous TV channels) are not correlated with the distribution of the next channel, so the model loss will not decrease in general. We provide an informal positive example in Appendix \ref{app:student}. 

Our definition makes no explicit mention of ``interestingness''. More precisely, interestingness is represented in our definition by the observer's choice of loss function $\ell$. Thus, for us, the interesting parts of artifacts are precisely those features which the observer decides are useful to learn about. Different observers can, and do, find different artifacts interesting, by virtue of the different parts of the feature space they choose to learn with their statistical model.

We hope that our definition will serve as a useful grounding for future work. On the theoretical side, it provides a basis for proving whether a system is open-ended. On a practical note, it raises the prospect of searching for open-ended systems. In this paper, we shall use it to underpin the argument that open-endedness lies on the critical path towards ASI, and in particular that the combination of open-ended algorithms and foundation models is ripe to yield significant progress towards that aim. We examine some subtleties of our definition in Appendix \ref{app:subtleties}.

\subsection{Related Definitions}

In the interests of space, we review the definitions of open-endedness most closely related to ours, covering more distantly related work in Appendix \ref{app:related_work}. \citet{chromaria} provided four necessary conditions for an evolutionary process to be open-ended, namely (1) that individuals must meet a minimal criterion in order to reproduce, (2) that evolution of individuals should create novel opportunities to meet the minimal criterion, (3) that individuals themselves should make decisions about how to interact with the world, and (4) that the potential complexity of the phenotype should not be limited by its representation. Our definition overlaps with these necessary conditions, but relaxes the constraint that the open-ended system is evolutionary. Our requirement that learnability is increasing can be seen as a generalisation of the minimal criterion in condition (1). Our requirement that the observer cannot intervene on the system is analogous to condition (3). Our requirement that novelty is increasing is analogous to conditions (2) and (4). Indeed, conditions (2) and (4) suggest that an open-ended system cannot be learned from a fixed data distribution.

To our knowledge, the most recent paper offering a definition of open-endedness is \citet{sigaud2023definition}. The authors write: ``an observer considers a process as open-ended if, for
any time $t$, there exists a time $t' > t$ at which the process generates a
token that is new according to this observer's perspective''. This definition has considerable overlap with ours. Like us, \citeauthor{sigaud2023definition} define open-endedness with respect to an observer. They consider the observer examining a sequence of tokens from a process, while we equivalently have the observer consider a sequence of artifacts from a system. Our requirement of novelty and learnability is compatible with their statement that the process should generate a token that is ``new according to the observer's perspective''. Our definition differs by being more precise about what this phrase means. In particular, we specify that what an observer considers ``new'' should be artifacts that are unpredictable according to their current statistical model of the system under consideration. Moreover, we specify that the observer's ``perspective'' is generated by learning that statistical model on the history of artifacts thus far presented by the system. 
In particular, our definition can rule out systems that display continual ``novelty'' but are otherwise uninteresting, like white noise on a TV screen, for instance. 

\subsection{Types of Observer}\label{sec:observers}

The choice of observer is a free parameter of great importance for our definition. From the perspective of AI research, there is a pre-eminent class of observers, namely humans. In other words, we wish to generate artifacts that are valuable to individual humans and to society. This provides a level of grounding for the open-ended system which narrows the search space considerably, as we shall argue in Section \ref{sec:oe-fm}. Nevertheless, our definition deliberately admits arbitrary observers, for several reasons. Firstly, it allows our definition to encompass open-ended systems which are not anthropocentric, such as biological evolution. Secondly, it allows us to reason about open-ended systems which might exceed human capabilities, so-called ASI. Thirdly, it allows us to determine whether systems can be open-ended with respect to any observer, as we did with the noisy TV.\footnote{There is one constraint on an observer which must be adhered to for our definition to make sense. The loss function must treat artifacts $X$ and predictions $\hat{X}$ on an equal footing. In particular it must be fixed in advance without any knowledge of the system $S$. Otherwise, an observer $O$ could find a system $S$ to be open-ended purely by discarding the artifacts from $S$ and constructing its own artifacts that it finds to be both novel and learnable.}

Practically speaking, any given observer will have some \textit{time horizon} $\tau$ which bounds their observations of a system, i.e. $t, T < \tau$. This concept allows us to distinguish between systems which are open-ended on different timescales. We say that a system is \textit{infinitely} open-ended with respect to an observer $O$ if it remains open-ended on any timescale $\tau \to \infty$. We say that a system is \textit{finitely} open-ended with time horizon $\tau$ with respect to an observer $O$ if it is open-ended for $t, T < \tau$. Consider, for example, an agent trained in simulation with an automatic curriculum over tasks. In principle, a human observer might find observations of the agent behaviour to be infinitely open-ended, for the agent may accrue the ability to solve ever more diverse and surprising tasks. In practice~\citep[cf. AdA,][]{ada}, novelty starts to plateau after about $1$ month of training, due to limitations in the richness of the task space and in the size of the agent's neural network. Thus AdA is finitely open-ended with time horizon $\approx 1$ month.

Similarly, an observer's judgement will be influenced by the limitations of their cognitive abilities relative to the breadth of the domain. For example, a human observer who reads a curriculum of ever more complex articles from a current snapshot of Wikipedia may find such a system open-ended, but only until they reach the limit of their memory. A suitable ordering of Wikipedia articles will present novel information, in the sense that every now and then an article will be more unpredictable than we have hitherto seen. We might also expect that this information will be learnable, because human knowledge is interlinked, in the sense that knowing more about one topic makes it easier to understand other topics that may crop up later. However, once human memory capacity is saturated, the human observer will start to forget previous articles. This violates learnability: in calculus, for instance, once one has forgotten the definition of a derivative, one will find it harder to understand an article about the chain rule. Therefore, conditioning on a history longer than an observer's recall doesn't necessarily make the current artifact more predictable. 

This example brings to light three interesting threads. Firstly, the open-endedness of human technology, as observed by humans, relies on our ability to compress knowledge into a form that can be maintained within our collective memory: indeed, we present an alternative definition of open-endedness in the language of compression in Appendix \ref{app:definition}. Secondly, an artificial superhuman intelligence may have less stringent memory constraints than humans, and therefore may judge itself to be open-ended beyond the point at which humans assess it to be so, re-emphasising that human observers must be considered pre-eminent for the purposes of safety, as we explore further in Section \ref{sec:safety}. Thirdly, the open-endedness in this example is a function of the breadth of the domain. In a narrower domain, elliptic curve cryptography say, the set of relevant Wikipedia articles would be much smaller, so a human observer would find this open-ended only until they had understood every article, at which point novelty would be violated. Nevertheless, humans can, and frequently do, make new discoveries in narrow domains via experimentation and reasoning; amassing a vast, static trove of data is not the be all and end all of open-endedness.

\subsection{Examples}\label{sec:examples}

In this section, we discuss some popular systems that are open-ended but not general, or that are general but not open-ended, with respect to a human observer. This serves two purposes. Firstly, it demonstrates that our definition is not so restrictive as to rule out systems that are intuitively open-ended, and is not so loose as to include systems that intuitively lack open-endedness. Secondly, it motivates the benefits that foundation models can provide in addressing the limitations of current open-ended systems and vice versa. 

Our first archetypal open-ended system is \textit{AlphaGo} \citep{silver2016mastering}. Consider as artifacts the sequence of policies produced across training by AlphaGo. After sufficient training, AlphaGo produces policies which are novel to human expert players, in the sense that they play moves which would be low probability for human professionals but which nevertheless are winning against the best humans. Furthermore, humans can improve their win rate against AlphaGo by learning from AlphaGo's behavior \citep{shin2023superhuman}. Yet, AlphaGo keeps discovering new policies that can beat even a human who has learned from previous AlphaGo artifacts. Thus, so far as a human is concerned, AlphaGo is both novel and learnable. AlphaGo is just one representative from a class of open-ended algorithms that augment reinforcement learning with \textit{self-play} \citep{samuel1959some}, achieving or exceeding human-level play in Go, Chess, Shogi \citep{silver2017mastering}, StarCraft II \citep{vinyals2019grandmaster} Stratego \citep{perolat2022mastering}, DotA \citep{berner2019dota}, and Diplomacy \citep{meta2022human}. 

AlphaGo is an example of an open-ended system that achieves narrow superhuman intelligence \citep{morris2023levels}. This limits its utility: self-play of this kind cannot by itself help us to discover new science or technology that requires combining insight from disparate fields, or taking actions across a range of modalities, timescales and contexts. The constraints of the game rules make the search for novel and learnable artifacts tractable, and these artifacts are found to be novel and learnable by human observers largely because it was humans who invented the game. 

Our second archetypal open-ended system is \textit{AdA} \citep{ada,xland}. AdA is a large-scale agent that learns to solve tasks in an 3D-environment called XLand2. In XLand2 there are 25B possible task variants, corresponding to different world topologies and a variety of possible games within each world, that are prioritized for learning potential~\citep{jiang2021prioritized}. Checkpoints of the AdA agent across training are open-ended with respect to a human observer who attempts to predict what capabilities the agent might show. Across training, the agent gradually accumulates zero-shot and few-shot capabilities over an ever wider set of held-out environments, requiring ever more complex skills. Thus the human continually observes novel capabilities in the agent. Furthermore, the prioritization of task variants provides an interpretable ordering to the accumulation of skills in the agent, rendering this learnable by a human. AdA represents a wider class of open-ended algorithms driven by \textit{unsupervised environment design} \citep[UED,][]{dennis2020paired, justesen2018illuminating}, which establish an \textit{automatic curriculum} \citep{leibo2019autocurricula, baker2019emergent} of environments in the zone of proximal development for agent learning~\citep{vygotsky1978mind}. 

It is natural to ask whether AdA would continue to be judged as open-ended by a human observer should training be continued indefinitely. Results in~\citet{ada} suggest that novelty starts to plateau, implying that with an order of magnitude more compute AdA would almost certainly not be open-ended. Indeed, the authors show that both increasing the size of the agent and increasing the number of tasks allow the agent to generalize to a wider range of environments. Thus, in order for this system to be open-ended on longer timescales, one would need an even richer environment and an even more capable agent to sustain the agent-environment co-evolution inherent in UED. 

Our third archetypal open-ended system is \textit{POET} \citep{poet, enhanced_poet}. POET trains a population of agents, each of which is paired with an environment that is evolving over the course of training. These paired agent-environment artifacts are open-ended with respect to a human observer seeking to model the features of the environments that arise, or equivalently the skills the paired agents possess. A \textit{Quality Diversity} algorithm \citep[QD,][]{qdnature, Mouret2015IlluminatingSS} is deployed with respect to the environments, hunting for challenging problems that lead to diverging performance across the population. QD is an example of a wider class of open-ended algorithms, namely evolutionary algorithms, which we encounter again in Section \ref{sec:evo_algs}. 

Crucially, POET periodically transfers agents from one environment to another, which results in an empirical example of the stepping stone phenomenon \citep{stanley2015greatness}: agents can eventually solve incredibly challenging environments that are not possible to solve with direct optimization. As a result of training for billions of environment steps, POET produces a diverse population of highly capable specialist agents, which can solve novel environments that are created through coevolution with the population \citep{brant2017minimal}. Novelty arises because of the mutation operator in the QD algorithm, which yields new and unpredictable environments. Learnability arises because each mutation is small, so the past lineage of an environment is a good guide to its current features. Just as for AdA, the key limitation on open-endedness is the environment parameterization itself: eventually POET will plateau once the agent can solve all possible terrains.

Our final example is contemporary \textit{foundation models}. These are a negative example; they are not open-ended by our definition with respect to any observer who can model their training dataset. The justification for this follows immediately from our consideration of the noisy TV in Section \ref{sec:formal_definition}. Contemporary foundation models are typically trained on fixed datasets. If the distribution of this data is learnable, which it must be, for the foundation model learned it in the first place, then it cannot be endlessly novel, because eventually the observer will have modelled the epistemic uncertainty. As we saw in Section \ref{sec:observers}, foundation models may appear open-ended to human observers if the domain of enquiry is sufficiently broad, by virtue of the memory limitations of the human brain. However, if the focus is narrowed, for instance to tasks that require planning \citep{momennejad2024evaluating, pallagani2023understanding, valmeekam2023planning}, the limitations of the foundation model in generating novel, correct solutions are exposed. 

Since foundation models are periodically retrained on new data, including data generated by their own interactions with humans and the real world, one could argue that the data distribution is not really fixed. In some quarters, this kind of distributional shift is seen as an annoyance, even one which threatens ``model collapse'' \citep{shumailov2023model}. We flip this argument on its head, and contend that augmenting foundation models with open-endedness offers a path towards ASI. Similarly, the fact that foundation models are typically conditional on context breaks the logic that they cannot be open-ended. In principle, the context of a foundation model can be recruited to recombine concepts in an open-ended way by leveraging some external measure of validity. This brings us neatly to some concrete suggestions for how to build open-ended foundation models.

\section{Open-Ended Foundation Models}\label{sec:oe-fm}

We have defined open-endedness and discussed why the
current foundation model training paradigm is \textit{not} open-ended. We believe that the trend of improving foundation models trained on passive data by scaling alone will soon plateau, and it will not be enough to reach ASI. Our position is that open-endedness is a property of any ASI, and that
foundation models provide the missing ingredient required for domain-general open-endedness. Further, we believe that there may be only a few remaining steps required to achieve open-endedness with foundation models. In the following subsections, we sketch four overlapping paths towards open-ended foundation models that lend credence to this belief. The paths are neither intended to be prescriptive nor exhaustive. Indeed, recent publications such as \citep{wong2023roles, sharma2023assembly} point to other paths. 

Before proceeding, we must justify our claim that a future foundation model trained passively on some large corpus of human data is unlikely to spontaneously acquire open-endedness. In principle, should we reach ASI, there will be some sum total of data which the model has consumed during its training, possibly via several intermediate stages. Therefore, our claim is not about the impossibility of assembling such a dataset. Rather, we suggest that it is unlikely that this dataset can be pre-collected offline in an efficient way. The reason is that open-endedness is fundamentally an \textit{experiential} process: producing novelty and learnability in the eyes of an observer requires continual online adaptation on the basis of the artifacts already produced, in the context of that observer's evolving prior beliefs. 

What would it take to collect offline a static dataset from which such an experiential skill could be learned? Such a dataset must contain a treasure trove of artifacts which themselves crisply show novelty and learnability. Yet the process by which culture evolves, ideas develop, inventions arise and technologies proliferate is seldom recorded neatly and comprehensively. The alternative paradigm, in which experience is ``built in'' to the open-ended system, is well illustrated by the scientific method. Since the Enlightenment, the simple process of making hypotheses on the basis of current knowledge, falsifying them with experiments based on a source of evidence, and codifying the results into new knowledge has yielded unprecedented progress in science and technology \citep{deutsch2011beginning}. In our view, the fastest path to ASI will take inspiration from the scientific method, compiling a dataset online by the explicit combination of foundation models and open-ended algorithms. 

\subsection{Reinforcement Learning}

The framework of Reinforcement Learning (RL) has been at the forefront of achieving superhuman performance in narrow domains, such as AlphaGo's groundbreaking strategies that have enriched the human understanding of the game of Go. RL agents act deliberately so as to shape their stream of experience for both accumulating reward (exploitation) and learning about how to increase expected reward in the future (exploration). A nuanced extension are agents that set their own goals to (learn to) pursue; and generating the sequence of these goals can itself be an open-ended process, which drives open-ended experience generation \citep{colas2022autotelic}. Voyager~\citep{wang2023voyager} provides an early example of how RL-like self-improvement can be built on top of foundation models, without the need for explicit parameter updates or established RL algorithms. Instead, Voyager assembles an LLM-powered curriculum, uses iterative prompting as an improvement operator, and assembles verified skills into a library for hierarchical reuse. 

A key problem in RL is how to shape exploration towards novel and learnable behaviors in high-dimensional domains, as discussed in \citet{jiang2022general}. Exploration can be guided, for instance, by pseudo-rewards \citep{bellemare2016unifying, burda2018exploration, du2023can}, modulation \citep{schaul2019adapting} or an automated curriculum that selects relevant tasks \citep{jiang2021prioritized,accel2022,samvelyan2023maestro}. To generalize this, a useful abstraction may be the notion of a \emph{proxy observer}, which sits within the system and proactively guides it to generate novel and learnable content for the true external observer. In the past this guidance was provided on the basis of simple metrics such as TD-error, but now we can leverage foundation models to guide exploration towards artifacts that more closely align with what a human observer deems to be novel and interesting~\citep{jiang2022general}. There is already evidence that this approach may be effective, with LLMs providing agent rewards from text in an environment \citep{klissarov2023motif} and compiling a curriculum of tasks based on their interestingness \citep{zhang2023omni, faldor2024omni}.

While RL considers the first-person perspective of an agent interacting with an environment, a different perspective centers on multi-agent dynamics, and the additional richness arising from all the ways that different (possibly heterogeneous) agents can interact with each other, adapt to each other, or learn from each other. The presence of multiple learning agents provides a source of non-stationarity, such that the optimal strategy for each individual will change over time, potentially in an open-ended manner. Non-stationary dynamics been used to achieve or exceed human-level performance in games like StarCraft, DotA and Stratego. There is early evidence that multi-agent systems may help to improve factuality and reasoning in LLMs via debate \citep{du2023improving,tang2023medagents}, although there is much more research needed before superhuman capability is reached.

\subsection{Self-Improvement}

To achieve open-endedness, a model must not only consume knowledge from pre-collected feedback as in, for example, RLHF \citep{ziegler2019fine}, but also generate new knowledge, in form of hypotheses, insights or creative outputs beyond the human curated training data. A self-improvement loop should allow the agent to actively engage in tasks that push the boundary of its knowledge and capabilities, for example via leveraging tools such as search engines, simulated environments, calculators or interpreters and interacting with other agents \citep{jiang2022general, schick2024toolformer}. This requires the model to have a scalable mechanism to evaluate its own performance, identify areas for improvement, and adapt its learning process accordingly.

There is growing evidence that foundation models can be leveraged for feedback in place of humans, and can significantly amplify data generated by humans. Examples include self-critique and revision for training harmless assistants \citep{bai2022constitutional} and guiding human evaluators \citep{saunders2022self}, self-correction for tool-use \citep{gou2023critic}, self-instruction for instruction following \citep{wang2022self}, self-debugging for code generation \citep{chen2023teaching}, self-rewarding for instruction following \citep{yuan2024self}, and leveraging VLMs as reward functions for control \citep{baumli2023vision}. These works hint at the possibility of foundation models generating their own samples and refining them in an open-ended way.

\subsection{Task Generation}

Closely related to both RL and self-improvement is the problem of task generation, also known as the ``problem problem'' \citep{leibo2019autocurricula}. One great candidate approach for open-endedness is to keep adapting the difficulty of tasks to an agent's capability so that they remain forever challenging yet learnable. Past examples of this type of system include setter-solvers \citep{schmidhuber1991possibility} and unsupervised environment design \citep{dennis2020paired, justesen2018illuminating, poet}. With the advent of foundation models, it has become feasible to use the Internet itself as an environment \citep{jiang2022general, gur2021environment} via web-based APIs, affording agents with an incredibly rich, ever-growing and human-relevant task domain \citep{zhou2023webarena}.

Another possibility is to instead learn world models---predictive simulators that can generate future outputs conditioned on text or actions. A promising approach is to consider a foundation model to be a world model itself, since it is capable of predicting the future \citep{wong2023word,gurnee2023language,park2023generative}. Learned world models like Genie \citep{bruce2024genie}, and text-to-video generation models like Sora \citep{videoworldsimulators2024} demonstrate that foundation video models can be used as learned simulators, including in real-world settings like robotics \citep{yang2023learning} and autonomous driving \citep{hu2023gaia1}. If these works combine with learned multi-modal reward models \citep{chan2023visionlanguage, du2023rewards}, they could be used to generate an open-ended curriculum of tasks, scaling to task spaces far larger and more photorealistic than can currently be achieved. At sufficient scale, this may provide a path to generating AI agents with superhuman adaptability across a wide range of previously unseen tasks, which can be deployed in the real world across the rapidly closing Sim-to-Real gap \citep{huang2023went}.

\subsection{Evolutionary Algorithms}\label{sec:evo_algs}

Evolutionary methods offer a promising path to generate open-ended systems with foundation models \citep{wu2024evolutionary}. LLMs are well-placed to act as selection and mutation operators, as they have been trained on vast datasets of human knowledge, culture and preferences. For example, LLMs offer a mechanism through which to make semantically meaningful mutations via text \citep{lehman2022evolution, meyerson2023language, chen2023evoprompting}. The simplest such approach may be via \emph{prompts}, which already allow foundation models to further improve their performance. Recent works have shown it is possible to far surpass human designed prompts, leading to stronger models \citep{fernando2023promptbreeder, yang2023large, guo2023connecting}. More recently, \citet{bradley2023qualitydiversity} and \citet{ samvelyan2024rainbow} went further, using an evolutionary algorithm and LLMs to both generate variation and evaluate the quality and diversity of candidate text, making it possible to guide the search for creative and novel outputs. In the future it may be possible to further refine a model on these outputs, or use them for planning \citep{gandhi2023strategic}, to achieve self-improvement.

Another angle of attack for evolutionary methods is in the space of code (also known as genetic programming). Foundation models have proven to be competent at producing diverse and novel programs, providing a means of iterating upon an archive of candidate solutions. For example, Eureka~\citep{ma2023eureka} evolves code-based reward functions to learn complex control behaviors. Similarly, FunSearch~\citep{romera-paredes2024mathematical} evolves programs that represent new mathematical knowledge. These examples are focused on specific domains, and it remains an open problem to scale code evolution to a more general setting.

\section{Achieving ASI Responsibly}\label{sec:safety}

Now that we have foundation models, designing a truly general open-ended learning system may be within our grasp. However, the power of open-endedness comes with a swathe of notable safety risks---beyond existing safety considerations facing foundation models~\cite{ecoffet2020open}. Finding solutions to these challenges are interesting and important core problems in open-endedness research. Because the solutions to these problems may well depend on the design of the open-ended system, it is critical that safety and open-endedness are pursued in tandem. We cover them here not to hold them separate from other directions in open-endedness---in fact many of these problems are current practical limitations of artificial open-ended systems. Rather, this section is intended to draw specific attention to these problems as some of the most fundamental and exciting directions for research in the field. Of course, this short section cannot do justice to the breadth of concerns. Hence, where possible, we provide references to the wealth of knowledge in the ASI safety community.

\begin{figure}[t!]
    \centering
    \includegraphics[width=\linewidth]{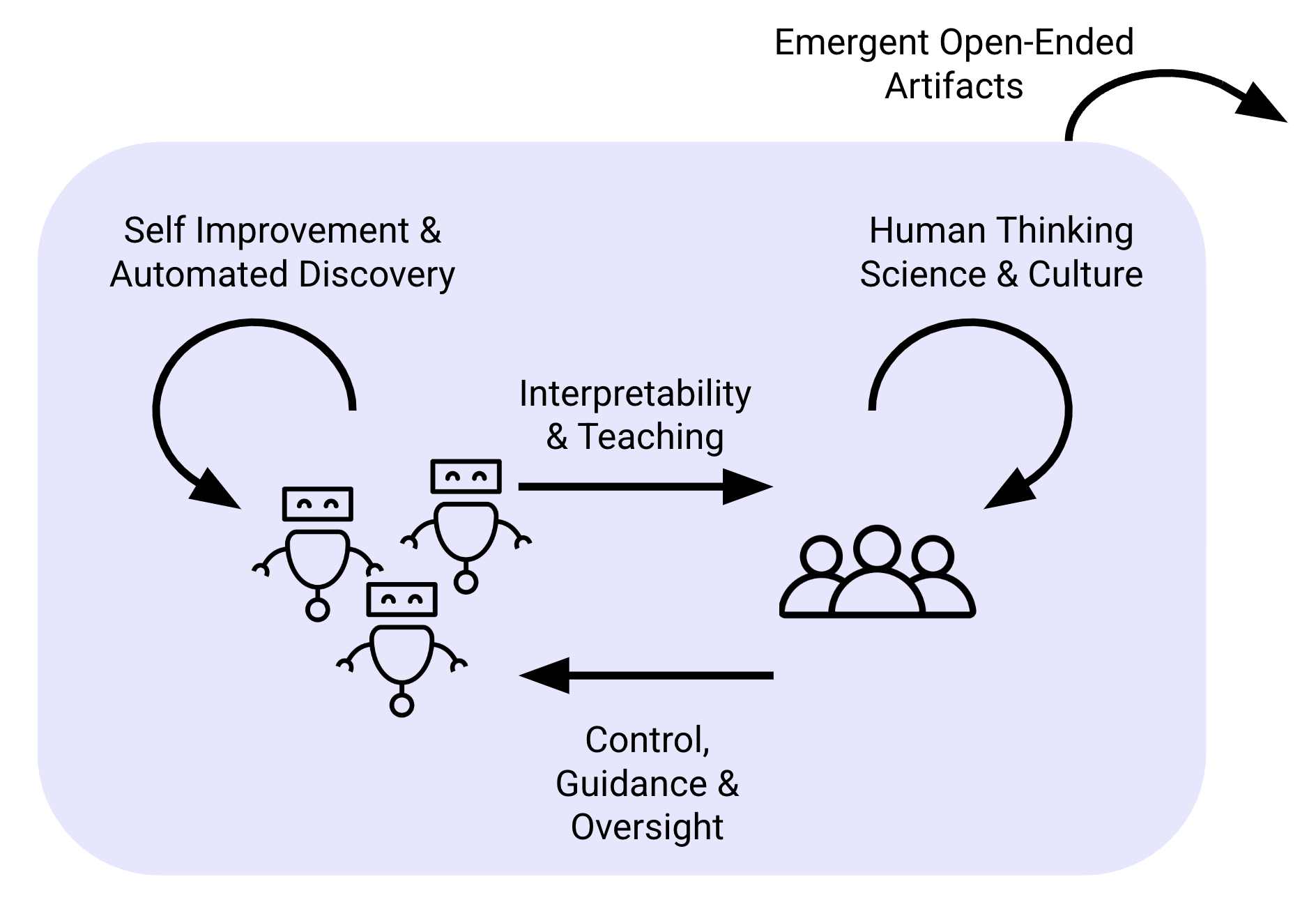}
    \vspace{-6mm}
    \caption{\textbf{Knowledge accumulation and transfer in a human-AI open-ended system.} We depict AI building on AI knowledge, humans understanding AI knowledge, AI understanding human knowledge, humans building on human knowledge, and emergent knowledge created by the process as a whole. Every process in this diagram offers an opportunity to embed safety methods that guide the system towards achieving ASI responsibly.
    }
    \label{fig:AIS_diagram}
    \vspace{-12pt}
\end{figure}

We organize our understanding of these risks similar to \citep{critch2020ai} by focusing on the ways knowledge is created and transmitted through the joint human-AI open-ended process in Figure \ref{fig:AIS_diagram}. A powerful open-ended system which has the problems listed in this section is not a beneficial open-ended system, and we believe it is not one we should be striving to build. Solving these problems is not just making open-ended systems safer, but also making them usable by humans. As such, addressing these problems should be thought of as minimum specifications of an open-ended system that we would want to build.

\subsection{AI Creation and Agency}
AI systems powering the open-ended creation of new knowledge could lead to powerful new affordances. Without direction, these creations could be the source of dual-use dangers~\cite{urbina2022dual}.
The danger is magnified when the open-ended systems take immediate action in an environment. Current state-of-the-art systems operate in narrow, simulated environments \citep{wang2023voyager, xland, ada}. However, as AI is trained in broader, more diverse simulations or is even deployed (and continues to learn) in the real world, it becomes critical to understand the dangers. The agency of open-ended AI poses several safety risks, such as goal misgeneralization~\citep{di2022goal,shah2022goal} and specification gaming~\citep{clark2016faulty}. Open-ended search can be seen as an ambitiously aggressive form of exploration; thus one could hope to use similar approaches to mitigate the dangers of exploration as in RL, like safe exploration~\citep{garcia2015comprehensive} and impact regularization~\citep{krakovna2018penalizing,turner2020conservative}.

\subsection{Humans Understanding AI Creations}
In order to provide informed oversight and direction when guiding an open-ended system, human observers need to at least partially understand the significance of the new artifacts that the system produces. This becomes increasingly challenging as the complexity of these artifacts grows, leading to the inability to give informed oversight and guidance. Such a system may not only be unsafe, but would no longer be open-ended for human observers, since it would no longer be learnable. As such, any open-ended system we want to build should have the ability to bring human observers along with it---understanding and interpreting these systems is not only a core problem to make them safe, it is also a core problem to make them useful.

One approach would be to try to understand the policy generated by open-ended systems through interpretability. With current approaches this would require a formidable interpretability effort for each domain of interest. However, with the advent of automated interpretability \citep{bills2023language}, one may hope to build increasingly good explanations of the systems' behaviors which match the increasing complexity of the open-ended system. This presents an sizeable challenge, as such a system would be a universal explainer~\citep{deutsch2011beginning}, by definition.

An alternative approach is to prefer designs for open-ended systems which promote interpretability and explainability, or whose goal is to teach human observers. Already, there are efforts to train systems which directly inform the user of implicit knowledge \cite{christiano2021eliciting}. One might aim to design systems that at least maintain informed oversight \cite{amodei2016concrete,bowman2022measuring}. This approach may be especially effective if the design of the open-ended system automatically facilitates understanding and control by human users~\cite{irving2018ai}.

\subsection{Humans Guiding AI Creation}
Even if we assume that human observers can understand enough of the behavior of an open-ended system to be in a position to give informed feedback, we arrive at the question of how a human designer could meaningfully guide an open-ended system. This challenge goes beyond the difficulties of directing individual RL agents, as not only do open-ended systems often lack well-defined objectives that could be modified, but they are increasingly unpredictable by design. One possibility would be to use humans in the loop to drive open-endedness~\citep{secretan2008picbreeder}, a kind of open-endedness from human feedback~\citep{zhang2023omni}. A complete solution to this problem not only needs to be directable, but must actively raise unexpected and possibly important artifacts to the user's attention.

If open-ended systems could be made as directable as individual RL agents, then work defining objectives which preserve controllability \cite{hadfield2016cooperative,hadfield2017off,carey2023human} might be a promising path towards more controllable open-ended systems. However, directing an open-ended system towards any objective effectively while maintaining the open-endedness is an open problem. This problem is not only important for safety, but is important for open-ended systems to be useful. In sufficiently broad domains---such as all of mathematics, all proteins, or all behaviors on a computer---an open-ended system may rabbit-hole into the obscure theorems, useless proteins, or only certain computer applications. Thus, building mechanisms that allow us to direct open-ended systems to not just the safe artifacts, but the interesting and useful artifacts, is a fruitful avenue for collaboration between safety and open-endedness researchers. 

\subsection{Human Society Adapting}

There are significant non-technical concerns in ensuring that society can understand, prepare for, and appropriately react to new technological capabilities emerging from open-ended foundation models. Indeed, the impact of AI systems is not just felt at the individual level, but also at the level of the collectives that structure our society---communities, organisations, markets and nation states, to name a few. Since the artifacts arising from open-ended foundation models will by definition appear novel, we must devote prospective attention to the ways in which these could harm or benefit the cooperative infrastructure of society \citep{dafoe2020open}. Likewise, we must develop mechanisms to avoid tipping points driven by feedback loops, like flash crashes \citep{aldrich2017flash}. Decision-makers should be prepared to adapt governance rapidly and retrospectively in response to open-ended artifacts, finding a good balance between collecting information and avoiding entrenchment of undesirable artifacts \citep{collingridge1980social}. 

\subsection{Emergent Risks of Open-Ended Systems}
Even if each subcomponent of Figure \ref{fig:AIS_diagram} can be made safe, it may still be the case that the aggregate joint human-AI open-ended system leads to unforeseen problems. 
For instance, two systems that are open-ended in isolation could negatively interact to cause neither to be open-ended. This would mean a cessation of progress and an inability to collectively respond to new challenges. While such emergent effects have been studied in multi-agent systems~\citep{johanson2022emergent} and ASI safety~\citep{critch2020ai} solutions are still elusive, and an understanding of these effects is critical to the safe deployment of open-ended systems. 

If such problems are inevitable and unpredictable, we would need our human-AI open-ended systems to adapt to solve novel ASI safety failures as they arise. Due to the inherent unpredictability of knowledge creation, these problems may be both unavoidable and solvable once as they arise~\citep{deutsch2011beginning}. We should be building an open-ended system whose safety is anti-fragile~\citep{taleb2014antifragile}, adapting to emerging safety risks and getting stronger for it. This entails designing techniques for understanding, monitoring, and rapidly coordinating responses to emerging risks.

\section{Conclusion and Outlook}
Foundation models have led to a rapid increase in the generality of current AI systems. However, current foundation models are limited in their capability to discover new knowledge. In this paper, our position is that to further advance in levels of AGI towards ASI, we require systems that are \emph{open-ended}---endowed with the ability to generate novel and learnable artifacts for a human observer. There has never been a more exciting time to build such systems, with foundation models already exhibiting general human-like knowledge that both accelerates further learning and guides this learning towards human-relevant artifacts. 

As we develop and deploy more generally-capable open-ended systems, novel safety concerns arise that will be critical to address. In order to realise the benefits of such systems, it is important that the human observer remains able to learn from the novel artifacts, bringing fields such as explainability to the forefront of open-endedness research. If these endeavors are successful, then we believe open-ended foundation models could lead to advances that drastically enhance modern society.

\section*{Impact Statement}

Our work provides a formal definition of open-endedness, and provides a discussion on its significance for the pursuit of ASI. We explore current research directions in the field, emphasising the potential of combining open-endedness with foundation models as a pre-eminent path towards achieving ASI. Developed responsibly, we believe that such open-ended foundation models can have tremendous positive impact on the society, accelerating scientific and technological breakthroughs, enhancing human creativity through a collaborative feedback loop, and acting as an engine for general knowledge expansion across many fields. Recognising the profound implications of this concept, we dedicate the entirety of \Cref{sec:safety} to an initial analysis of potential risks and societal impacts, offering frameworks for the responsible and ethical development of ASI. We hope that highlighting these issues early will help to promote safety, responsibility and accountability as the field grows.

\section*{Acknowledgements}
We gratefully acknowledge Dave Abel for providing valuable feedback on an early draft of this paper. We are thankful to the designers at the \href{thenounproject.com}{Noun Project}, from which we sourced graphics under the CC BY 3.0 licence as follows: ``\href{https://thenounproject.com/browse/icons/term/tick?iconspage=1}{tick}'' icon by kareemovic, ``\href{https://thenounproject.com/browse/icons/term/Delete?iconspage=1}{Delete}'' icon by kareemovic, ``\href{https://thenounproject.com/browse/icons/term/alien?iconspage=1}{alien}'' icon by Artem Yurov, ``\href{https://thenounproject.com/browse/icons/term/girl?iconspage=1}{girl}'' icon by Teewara soontorn, ``\href{https://thenounproject.com/browse/icons/term/year-of-rat?iconspage=1}{year of rat}'' icon by DailyPM, ``\href{https://thenounproject.com/browse/icons/term/aircraft?iconspage=1}{aircraft}'' icon by mikicon, ``\href{https://thenounproject.com/browse/icons/term/concorde?iconspage=1}{concorde}'' icon by mikicon, ``\href{https://thenounproject.com/browse/icons/term/plane?iconspage=1}{Plane}'' icon by CAMB, ``\href{https://thenounproject.com/browse/icons/term/humans?iconspage=1}{humans}'' icon by Ifanicon, and ``\href{https://thenounproject.com/browse/icons/term/robot?iconspage=1}{Robot}'' icon by Deemak Daksina.

\bibliographystyle{abbrvnat}

\bibliography{main.bib,zotero.bib}

\newpage
\appendix

\section{Illustrating Open-Endedness}

\subsection{An Informal Example}\label{app:student}

To illustrate our definition informally, we provide a relatable real-world example. Let $S$ be a research lab and the $x_t$ be academic papers published by the lab. A natural choice of observer $O$ is a research student in the field at a different lab. Roughly speaking, a research student sees novelty in a line of work if, based on their knowledge of the literature up to time $t$, given any subsequent paper $x_T$ they can always find a later paper $x_{T'}$ that is more surprising than $x_T$. This is intuitively sensible, a putative student with knowledge of Newtonian mechanics will find Maxwell's equations hard to predict, quantum mechanics even more surprising, and contemporary particle physics very far outside their current level of comprehension. A research student sees learnability in a line of work if they find that reading the previous papers helps them better to predict the contents of the current paper. Again, this appeals to our intuition: part of the purpose of citations, for instance, is to point new researchers at previous works that will help to deepen their understanding of the current work. 

Our interpretation of ``interestingness'' as learnability also makes sense from the perspective of a research student. A research student may choose to ignore a paper's choice of font, but will likely pay close attention to the details of a novel method that yields state-of-the-art results. Thus the student finds interesting the parts of the paper from which they can learn the most. Similarly, the requirement that the loss metric $\ell$ be chosen without knowledge of $S$ finds a natural interpretation here. A research student cannot judge the open-endedness of a stack of papers by choosing to never read the papers and instead inventing their own line of research with no reference to previous works. 

\subsection{Definitional Subtleties}\label{app:subtleties}

Self-play illustrates some subtleties in our definition. The first subtlety is the dependence of open-endedness on the choice of observer. Suppose that $O$ is an oracle who knows the Nash strategy to play in Go. Assuming that the oracle is modelling the win-rate of AlphaZero's artifacts against its own policy, it will never find any AlphaZero policy to be novel. Therefore the oracle does not find AlphaZero to be open-ended. The second subtlety is the dependence of open-endedness on the learning limitations of the observer. To an average human Go player, as opposed to an expert, AlphaZero becomes novel earlier in training, and at some point ceases to be learnable, because the average player cannot figure out how to improve their own play with reference to very unusual style of a superhuman policy. Thus, open-ended systems only remain open-ended while they can ``educate'' their observers. We posit that superhuman intelligence will be interesting to humans only as far as humans can learn to understand it. The third subtlety is that open-ended systems need not explore a problem space fully to qualify as open-ended. Recently, adversarial search was shown to yield policies that beat reimplementations of AlphaZero and which are so simple that even amateur humans can learn them \citep{wang2023adversarial}. Novelty and learnability give no guarantee of coverage.

Because our definition is based on the perspective of an external observer, one could worry that this makes it impossible to make any sort of objective claims about the open-endedness of any particular system, in harmony with the arguments of \citet{stanley2016role,stepney2023open}. There are two factors which mitigate this concern. Firstly, the definition of open-endedness becomes objective given any fixed observer, and so it becomes a measurable claim, in the sense that theorems can be written and experiments conducted. For instance, if we care about open-endedness with respect to humans, open-endedness can be measured experimentally by how well humans can predict the system. By having observer-dependence explicit in our definition, we make precise the intuition that different observers, with different prior knowledge, different cognitive capabilities and different timescales, are likely to judge the same system in different ways. Thus our definition gracefully encompasses the diversity in perspectives of human individuals and groups (such as companies or governments), as well as the possibility that AI systems themselves could be observers.

Secondly, while our definition of open-endedness depends on an external observer, it is an open question as to whether all ``reasonable'' observers would judge the same systems to be open-ended. Since our definition rests on a notion of predictability with respect to the observer, our definition will be as subjective as the underlying notion of predictability. One may believe that predictability can be accurately and objectively modeled as Solomonoff induction \citep{solomonoff1960preliminary}. Thus if reasonable observers are taken to be those whose predictions eventually follow something approximating Solomonoff induction, then any observer in this class would eventually agree on which systems are open-ended.

Practically speaking, there are various existing methods in the literature which can immediately be adapted to assess the open-endedness of a system. First, one might elicit direct human feedback on learnability and novelty of artifacts, in the same spirit as RLHF \citep{ouyang2022training} or PicBreeder \citep{secretan2008picbreeder}. Second, one can use large language models themselves as judges of novelty and learnability, as argued for in OMNI \citep{zhang2023omni}. Finally, one could explicitly learn a model of the artifacts with an online learning method like Follow-the-Regularized-Leader \citep{hazan2010extracting}.

Can an open-ended system be its own observer? In principle, there is nothing in our definition that rules out self-observing open-ended systems. For example, an individual self-improving agent could generate a series of artifacts, each one of which is novel (surprising compared to the previous artifacts) and learnable (increasingly predictable given the more history of the past artifacts). When the feedback from self-observation is used to improve the system itself, we call the observer a \textit{proxy observer} for it no longer sits outside the system. 

For example, AlphaGo can be seen as an example of a self-observing system, in that the agent trains in self-play i.e. it observes its own policy as an opponent, is challenged by the novel discoveries of search, and learns from them to improve the policy. Likewise, humans can experience ``Eureka moments'', when an individual suddenly reconceptualizes a problem in a ways that yields a solution \citep{sternberg1995nature}. A series of Eureka moments, each building on the last, is a self-observing open-ended system: the human generates discoveries which are novel to themselves, but which are also predictive of the next discovery.

Our notions of learnability is rather strict, in that it requires that the loss be decreasing for all $t' > t$. A weaker and more practical notion of learnability might state that it should be probabilistically unlikely that the loss will increase as a function of $t$: 
\begin{equation*}
  \forall T, \forall t < T, \forall T > t' > t : \mathbb{P} \left(\ell(t',T) \geq \ell(t,T)\right) <\delta \, .
\end{equation*}

It would be interesting to compare the consequences of $\delta$ being a constant with the situation in which $\delta$ has some appropriate dependence on the variables $(t,t',T)$. Similarly, one could weaken the notion of novelty to state that it should be probabilistically unlikely that the loss will decrease as a function of $T$. We believe that there may be several related and differently useful variants on our definition that would be interesting to independently study, in a similar way that there are many notions of convergence which are interesting, related, and differently useful.

\section{Alternative Definition}\label{app:definition}

In Section \ref{sec:formal_definition} we provided a formal definition of open-endedness in the language of statistical learning. Here we give an alternative definition which we conjecture is equivalent under appropriate conditions. The alternative definition is phrased in the language of compression, a topic with known formal connections to statistical learning \citep{hutter2004universal,david2016statistical,campi2023compression, deletang2023language}.

A \textbf{system} $S$ produces a sequence of \textbf{artifacts} $X_t \in \mathcal{X}$, indexed by time $t$. An \textbf{observer} $O$ processes a new artifact $X_T$ to determine its information content given a history $h_t = X_{1:t}$ of past ones. $O$ possesses a history-dependent compression map $C_{h_t}: \mathcal{X} \rightarrow \{0,1\}^*$ which encodes $X_T$ into a binary string of length $|C_{h_t}(X_T)|$.

The system displays \textbf{novelty} if the information content increases, namely:
\begin{equation*}
\forall t, \forall T > t, \exists T' > T: |C_{h_t}(X_{T'})| > |C_{h_t}(X_T)|.
\end{equation*}
In other words, the complexity of the artifacts grows, according to the observer.

The system is \textbf{learnable} if conditioning on a longer history increases compressibility, namely:
\begin{equation*}
\forall T, \forall t < T, \forall T>t'>t:|C_{h_{t'}}(X_T)| < |C_{h_t}(X_T)|.
\end{equation*}
In other words, as its history grows, the observer must be able to keep extracting additional patterns that help it compress future artifacts.

Finally, a system is \textbf{open-ended} from the perspective of $O$ if and only if it generates sequences of artifacts that are both novel and learnable.

\begin{figure}[t]
    \centering
    \includegraphics[width=\linewidth]{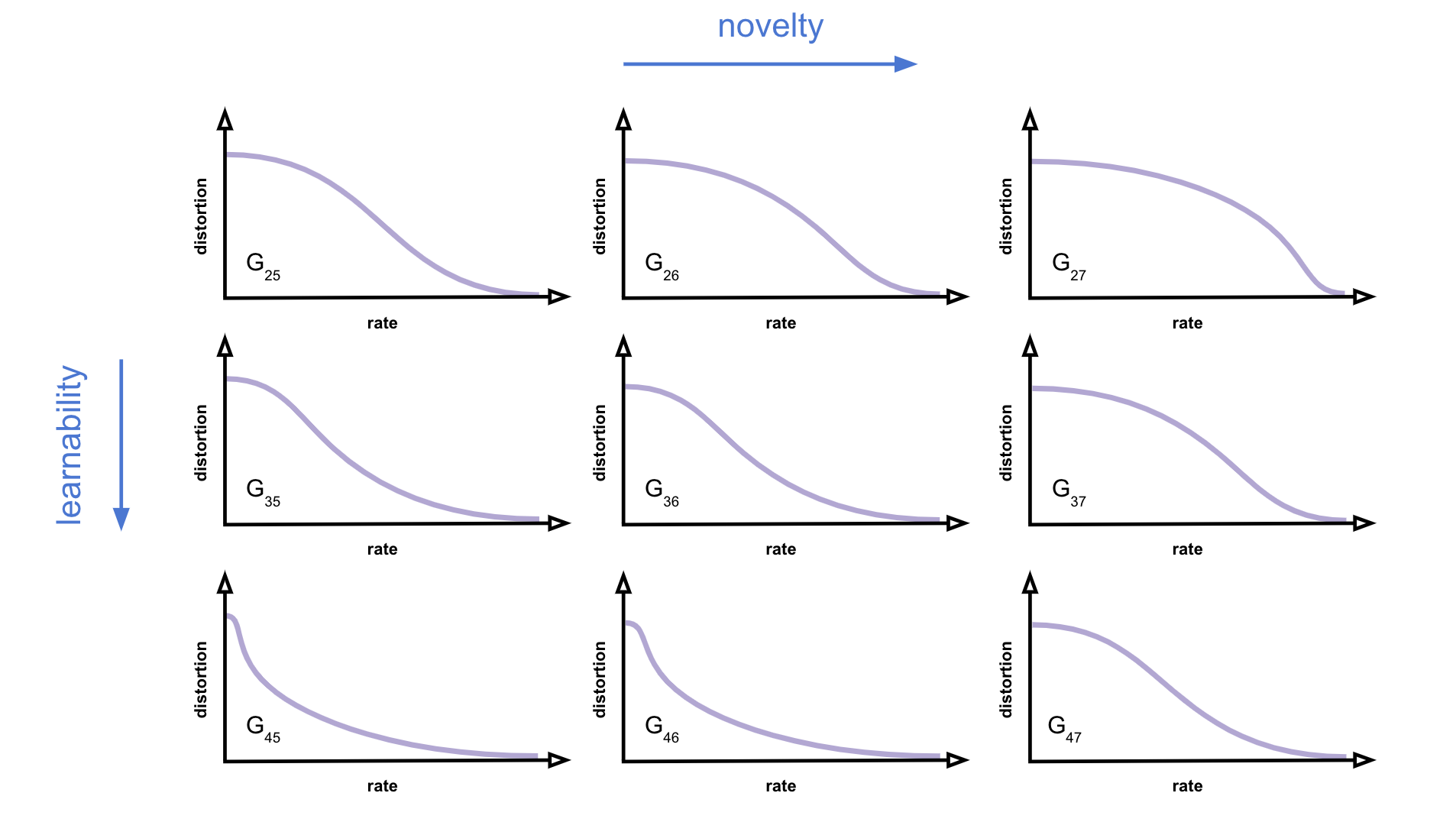}
    \caption{\textbf{Open-endedness through the lens of rate-distortion curves.} We depict part of the upper triangular matrix of rate-distortion curves $G_{tT}$ induced an observer after seeing the first $t$ artifacts aiming to lossily compress future artifact $T$. Here $t = 2,3,4$ and $T = 5,6,7$. Broad novelty is the property that, as you move from left to right in any fixed row, the rate-distortion curves become fatter. Broad learnability is the property that, as you move from top to bottom in any fixed column, the curves become flatter. For the system to be broadly open-ended, both properties must hold.}
    \label{fig:rate_distortion}
\end{figure}

We allow for the compression map $C_{h_t}$ to be \textbf{lossy}. Hence, $O$ also possesses a decompression map $D_{h_t}: \{0,1\}^* \rightarrow \mathcal{X} $, a symmetric loss function $\ell: \mathcal{X}\times\mathcal{X} \rightarrow \mathbb{R}^+$, and a threshold $\epsilon \in \mathbb{R}^+$ that upper-bounds the error made by by $C_{h_t}$:
\begin{equation*}
\forall T, \forall t < T: \ell(D_{h_t}(C_{h_t}(X_T)), X_T) < \epsilon.
\end{equation*}
We can strengthen the definition to be independent of $\epsilon$ by appealing to rate-distortion theory. A \textbf{rate-distortion} curve plots the the minimum information content $|C_{h}(X)|$ such that $\ell(D_{h}(C_{h}(X)), X) < \epsilon$ against $\epsilon$, where the minimum is over the maps $C_h$ and $D_h$. The information content is referred to as the rate and $\epsilon$ is referred to as the distortion. Picture a grid of rate-distortion curves $G_{tT}$ indexed by (discretized) $t$ and $T$, as in Figure \ref{fig:rate_distortion}. Remember that $T > t$, so $G_{tT}$ is strictly upper triangular, with other entries being undefined. Then \textbf{broad novelty} is the requirement that the curves get ``fatter'' as you move across the columns $T$ on the grid, for every row $t$. Similarly, \textbf{broad learnability} is the requirement that the curves get ``flatter'' as you move down the rows $t$ on the grid, for every column $T$. \textbf{Broad open-endedness} is the requirement that both broad novelty and broad learnability hold. This notion of broad open-endedness is vague in the same way the notion of ``convergence'' is vague in that it can be made precise in many subtly different but connected ways. For instance, one could say a system is ``uniformly'' open-ended if distortion increases across the rows and decreases down the columns for every rate $\epsilon$. Alternatively, one could define ``average'' open-endedness by requiring that the integral of the rate-distortion curve get larger as you move across the columns and smaller as you move down the rows. We hope that future work will elucidate these subtleties in defining broad open-endedness and determine which variants have theoretical or practical merit.

\section{Further Related Work}\label{app:related_work}

Open-endedeness as a term emerged from the AI Life community when trying to quantify and replicate the increasing complexity and perpetual novelty of biological evolution. This is a rich field with a significant degree of disagreement \citep{earle2021video}.  As such there are a wide range of metrics proposed within the context of evolutionary systems which aim to quantify it's behavior. For instance  persistence filtering, which measures how many generations an organism has persisted for~\citep{dolson2019modes}, and evolutionary activity statistics~\citep{bedau1997comparison, bedau11998classification}. The closely related question around the necessary conditions to produce open-ended evolution has also been deeply studied~\citep{taylor2018routes,taylor2015requirements}. As these definitions are largely specific to biological evolution, we focus the remainder of our discussion on the more recent definitions which aim to define open-ended systems in a way that applies to current ML systems and systems more broadly.

Our definition of open-endedness is closely related to the concept of potential surprise in economics \citep{shackle1949expectation}. To measure \textit{potential surprise}, an individual should ask: ``how surprised would I be if this outcome actually occurred, if, at the time it occurred, I were still looking at the world in the way I look at it right now?'' \citep{derbyshire2017potential}. Interpreting surprise as unpredictability under a statistical model, an open-ended system $S$ is precisely one which produces ever increasing ``Shackle surprise'' in an observer which is learning. The concept of potential surprise is itself based on the century-old idea of Knightian uncertainty \citep{knight1921risk}. \textit{Knightian uncertainty} is a lack of any quantifiable knowledge about some possible occurrence, as opposed to the presence of quantifiable risk. Thus, somewhat imprecisely, an open-ended system $S$ is one which induces Knightian uncertainty in an observer who is learning. 

In \citet{stanley2015greatness}, the authors argue that local search for novel and interesting artifacts can be advantageous over optimization for a global objective. This is because stepping stones towards a solution that optimizes the global objective may well not resemble the solution itself. Hence it is hard to translate the global objective into a local improvement operator that reliably accumulates improvements without getting stuck in local optima. To address this deceptiveness, they suggest that novelty search \citep{lehman2011abandoning}, guided by a notion of interestingness, can uncover stepping stones that advance knowledge and capability. We take inspiration from this blueprint and turn it into a definition. In order to clarify the notions of novelty and interestingness, we formalize them with respect to an external observer. Novelty becomes unpredictability according to the observer's history-conditional model, and interestingness becomes learnability of that model across the history of observations.

Our definition naturally relates to the notion of curiosity. Curiosity, implemented as prediction error of a world model, has long been mooted as an intrinsic motivation that can lead to open-ended discovery in RL agents given a sufficiently rich environment space \citep{schmidhuber1991possibility,pathak2017curiosity,raileanu2020ride,henaff2023exploration}. Our definition of novelty is effectively a generalisation of curiosity, without requiring an overarching RL framework. Our requirement of learnability ensures that the observer attempts to capture all the epistemic uncertainty about the artifacts produced by a system. One challenge is that curiosity based on novelty alone leads to ``stochastic traps'', whereby an agent will seek out sources of random noise with which to sate its curiosity \citep{schmidhuber1991adaptive, burda2018exploration, shyam2019model}. In principle, our definition of novelty collapses such aleatoric uncertainty by taking the expectation. In practice, we can only estimate the expectation, so it may be useful to subtract from the loss an estimate of the aleatoric uncertainty as in \citet{mavor2022stay}. We hope that future work will examine such subtleties required for an algorithmic implementation of our definition. 

The synergies between foundation models and open-endedness have previously been discussed by \citet{jiang2022general}. The authors propose a general notion of exploration and detail how open-endedness can be used to solve exploration problems when training foundation models. Our work follows in this line of thinking, providing a formal definition of open-endedness to make the  discussion precise, and further developing the connections between open-endedness and ASI. A construction of a particular open-ended learning system is provided in \citep{jiang2022general}, which may or may not fit our proposed definition of an open-ended system depending on how it is instantiated. The system generates Turing machine descriptions of MDPs, explicitly optimizing for an objective containing terms for learning potential, diversity, and grounding. These terms have some high-level relation to our notions of learnability and novelty, but they are quite distinct in the details. For instance, learning potential is divided into three sub-critia, improbability, learnability, and consistency, which are not made entirely formal. More crucially, the learnability discussed by \citep{jiang2022general} is a property of a single MDP, whereas the learnability we define is a property of a sequence of artifacts. Similarly, in \citep{jiang2022general} diversity is defined as a distance measure between MDPs, whereas novelty, as we define it, is a property of the learning of the observer with no necessary relationship to distances in the space of artifacts. It would be an interesting direction for future research to understand under what conditions the system described in \citep{jiang2022general} would be open-ended by our definition, and, more generally, whether one can directly optimize for open-endedness in some circumstances.

Open-endedness is related to, but separate from, the notion of an AI-generating algorithm \citep[AIGA,][]{clune2020aigas}. An AIGA automatically learns how to build a general AI, based on meta-learning model architectures, meta-learning learning algorithms, and automatically generating data from which to learn. Adapting the logic of \citet{clune2020aigas}, an AIGA need not be open-ended by our definition; if an AIGA had the objective of passing a Turing test, it need not produce any further novelty once this objective had been achieved. Likewise, an open-ended system need not be an AIGA; as we shall see in Section \ref{sec:examples}, there exist open-ended systems with narrow scope that match or exceed human ability without full domain-generality. Our idea of an Open-Ended Foundation Model in Section \ref{sec:oe-fm} lives at the intersection between open-endedness and AIGAs. 

Similarly open-endedness is related to, but distinct from, continual RL \citep{abel2023definition}. A continual RL problem is one in which the best agents never stop learning. However, as observed by \citep{sigaud2023definition}, this does not necessarily imply that the agent policies \textit{accumulate} increasing novelty. Rather, a continual RL agent could cycle among some set of strategies. In the case where continual RL does produce policies which are open-ended according to some observer, this open-endedness will have a scope that is restricted by the environment. 

\end{document}